\documentclass[10pt,twocolumn,letterpaper]{article}

\usepackage{iccv}              
\usepackage[accsupp]{axessibility}
\usepackage{epsfig}
\usepackage{graphicx}
\usepackage{amsmath}
\usepackage{amssymb}
\usepackage[noend]{algpseudocode}
\usepackage{url}
\usepackage{graphicx}
\usepackage{enumitem}
\usepackage{caption}
\usepackage{xcolor}
\usepackage{algorithm}
\usepackage{mathtools}
\usepackage{color, colortbl}
\usepackage{url}

\usepackage{caption}
\usepackage{subcaption}
\usepackage{float}

\usepackage{multirow}
\usepackage{pifont}
\include{mathcommand}
\usepackage{amsmath,amsfonts,bm}

\def\eqref#1{equation~\ref{#1}}

\def\1{\bm{1}}

\def\eps{{\epsilon}}

\DeclareMathAlphabet{\mathsfit}{\encodingdefault}{\sfdefault}{m}{sl}
\SetMathAlphabet{\mathsfit}{bold}{\encodingdefault}{\sfdefault}{bx}{n}

\makeatletter
\DeclareRobustCommand\onedot{\futurelet\@let@token\@onedot}
\def\@onedot{\ifx\@let@token.\else.\null\fi\xspace}
\def\eg{\emph{e.g}\onedot} 
\def\ie{\emph{i.e}\onedot}

\def\etal{\emph{et al}\onedot}
\makeatother

\def\Vec#1{{\boldsymbol{#1}}}
\def\Mat#1{{\boldsymbol{#1}}}
\definecolor{Lightgray}{gray}{0.95} 
\definecolor{blond}{rgb}{0.98, 0.94, 0.75}
\definecolor{lightblue}{rgb}{0.1, 0.1, 0.75}
\definecolor{aliceblue}{rgb}{0.94, 0.97, 1.0}
\definecolor{cornflowerblue}{rgb}{0.27, 0.51, 0.8}
\definecolor{iccvblue}{rgb}{0.21,0.49,0.74}
\usepackage[pagebackref,breaklinks,colorlinks,allcolors=iccvblue]{hyperref}

\def\paperID{3736} 
\def\confName{ICCV}
\def\confYear{2025}

\title{TITAN-Guide: \underline{T}aming \underline{I}nference-\underline{T}ime \underline{A}lig\underline{N}ment for Guided Text-to-Video Diffusion Models}

\author{%
  Christian Simon$^{\dagger}$ \quad Masato Ishii$^{\clubsuit}$ \quad Akio Hayakawa$^{\clubsuit}$ \quad Zhi Zhong$^{\dagger}$ \vspace{-0.06cm}\\ Shusuke Takahashi$^{\dagger}$  \quad Takashi Shibuya $^{\clubsuit}$ \quad Yuki Mitsufuji$^{\dagger, \clubsuit}$  \vspace{0.3cm} \\
  $^{\dagger}$Sony Group Corporation \quad $^{\clubsuit}$Sony AI  \vspace{-0.1cm} \\
  {\fontsize{9}{16}\texttt{\{first\_name.last\_name\}@sony.com}} \vspace{-0.01cm} 
  \\ 
}

\usepackage{multicol}
\usepackage{capt-of,etoolbox}


\begin{document}
\twocolumn[{%
\renewcommand\twocolumn[1][]{#1}%
\maketitle
\vspace{-1.3cm}
\begin{center}
    \centering
     \captionsetup{type=figure}
    {
     {\includegraphics[width=0.95\textwidth]{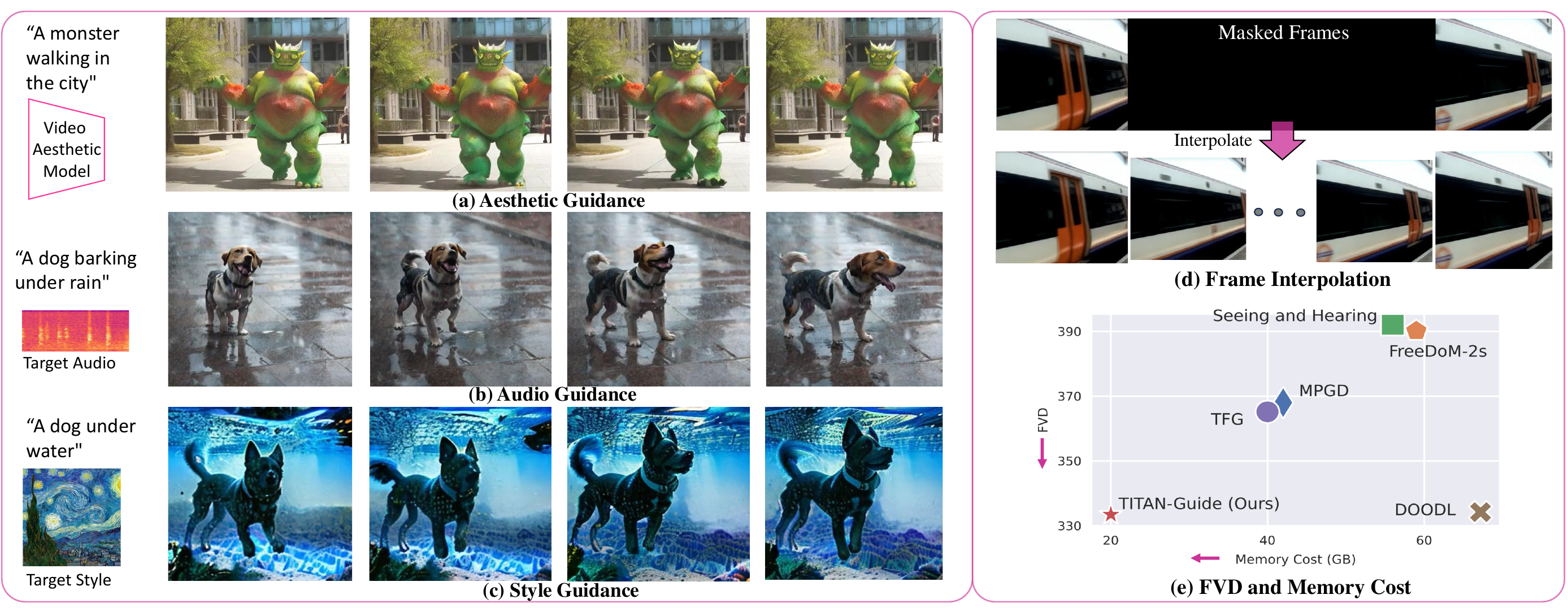}}
    }
    \captionof{figure}{ Various guided diffusion tasks with objectives to steer the output of generated videos and performance comparison with the state-of-the-art techniques. (a) A generated video by TITAN-Guide for aesthetic and composition tasks. (b) A generated result of our proposed approach on the multimodal alignment task between video and audio. (c) A style guidance task result of TITAN-Guide. (d) A result on frame interpolation tasks completing frames in the masked segments. (e)  Comparison between TITAN-Guide and state-of-the-art techniques in guided diffusion models with the label
Fréchet Video Distance (FVD, the lower the better) and GPU VRAM's memory cost for the audio guidance task on
VGG-Sound~\cite{chen20vggsound}.  }
\label{fig:teaser}
\end{center}%
}]



\vspace{-.2cm}
\begin{abstract}
In the recent development of conditional diffusion models still require heavy supervised fine-tuning for performing control on a category of tasks. Training-free conditioning via guidance with off-the-shelf models is a favorable alternative to avoid further fine-tuning on the base model. However, the existing training-free guidance frameworks either have heavy memory requirements or offer sub-optimal control due to rough estimation. These shortcomings limit the applicability to control diffusion models that require intense computation, such as Text-to-Video (T2V) diffusion models.  
 In this work, we propose Taming Inference Time Alignment for Guided Text-to-Video Diffusion Model, so-called TITAN-Guide, which overcomes memory space issues, and provides more optimal control in the guidance process compared to the counterparts. In particular, we develop an efficient method for optimizing diffusion latents without backpropagation from a discriminative guiding model. In particular, we study forward gradient descents for guided diffusion tasks with various options on directional directives. In our experiments, we demonstrate the effectiveness of our approach in efficiently managing memory during latent optimization, while previous methods fall short. Our proposed approach not only minimizes memory requirements but also significantly enhances T2V performance across a range of diffusion guidance benchmarks.  Code, models, and demo are available at \url{https://titanguide.github.io}. \vspace{-.8cm} 
\end{abstract}

\section{Introduction}




While pretrained diffusion models excel at Text-to-Video (T2V) tasks, challenges still arise when generating content on the basis of specific conditions. One key issue is ensuring proper guidance and adaptation of pretrained models for improved safety and alignment with specific requirements. Some emerging methods~\cite{meng2021sdedit,zhang2023controlnet,mou2024t2i,cong2023flatten} address guidance-free generation through model parameter updates, but these methods incur additional training costs. In contrast, \textit{training-free guidance} methods produce samples that align with specific targets using a differentiable target predictor. This eliminates the need for parameter fine-tuning, yielding a more efficient alternative.

To enable training-free guidance, classifier guidance~\cite{song2020classifierguidance} adjusts the diffusion model’s score estimate using the gradient with any differentiable loss function. Advancements in diffusion models using classifier guidance~\cite{xing24seeing,song2023ugd,he2024mpgd,ye2024tfg,yu2023freedom,wallace2023doodl,novack2024ditto} have achieved improved alignment in the image domain. Direct optimization with end-to-end  guidance ~\cite{wallace2023doodl} is promising to dramatically steer the generation in diffusion models.  
However, a key challenge with direct optimization persists due to the limited graphics processing unit (GPU) memory budget required for navigating multiple denoising steps. To address this, several diffusion-guidance methods~\cite{ye2024tfg,song2023ugd,he2024mpgd,yu2023freedom,nair2023steered} depend on the estimation of the clean-data distribution for efficiency, using a single denoising step, \ie, Tweedie's estimation~\cite{Efron2011TweediesFA}. However, in the early denoising steps, this estimation often generates incomplete visual outputs, limiting the classifier models' capability. Therefore, classifier guidance methods tend to misdirect the diffusion sampling process. These limitations restrict the applicability of existing guiding methods to work on Text-to-Video (T2V) diffusion models, which are memory intensive. As shown in Fig.~\ref{fig:teaser}, previous methods with different optimization strategies face trade-offs between memory usage and output quality.

To effectively guide the diffusion model while optimizing memory usage, we present our diffusion-guidance method Taming Inference-Time AligNment for Guided Text-to-Video Diffusion Model (TITAN-Guide). We also investigated a classifier guidance technique leveraging forward gradient descents~\cite{baydin2022fwdgrad} to reduce memory consumption while enhancing control over the generation process. 

\noindent The contributions of our study are three-fold:
\begin{itemize}[leftmargin=0.6cm]
    \item We propose TITAN-Guide, which has  efficient memory usage to guided T2V diffusion models.
    \item We investigated the forward gradient method for classifier-guidance diffusion models with various gradient approximation techniques.
    \item We adapted and extended text-to-image guidance techniques for video diffusion, setting a new benchmark in the field and advancing the capabilities of guided T2V models.
\end{itemize}

\section{Related Work}


\paragraph{Classifier guidance.} A classifier guidance method~\cite{song2020classifierguidance} builds upon a diffusion model by modifying the input noise vector at intermediate states, using an estimated latent representation derived from the Denoising Diffusion Implicit Model (DDIM)~\cite{song2020ddim}. Prior methods~\cite{ye2024tfg,yu2023freedom,he2024mpgd,bansal2023universal,meng2021sdedit,Nair2023steerdiff,xing24seeing} have successfully implemented guided diffusion models by incorporating different variations in the process of updating latents during the iterative steps. FreeDoM~\cite{yu2023freedom} introduces a novel \textit{time-travel strategy}, which operates by denoising 
  in a back-and-forth manner. FreeDoM emphasizes the critical need to adjust the guidance strength at various time steps to achieve optimal results. For high-quality and faster generation, He~\etal~\cite{he2024mpgd} leverage the manifold hypothesis, refining the precision of guided diffusion steps to accelerate the process. Ye~\etal~\cite{ye2024tfg} introduced a unified Training-Free Guidance (TFG) framework, integrating existing methods such as those of He~\etal~\cite{he2024mpgd} and Yu~\etal~\cite{yu2023freedom} as special cases. TFG offers a holistic perspective on training-free guidance, operating within an algorithm-agnostic design space. Through both theoretical and empirical analysis, TFG also presents an efficient and effective strategy for identifying optimal hyperparameters. While these methods demonstrate promising results in guiding diffusion models, their application to T2V models remains challenging due to the dependence on output estimation at each optimization step and the substantial memory requirements of backpropagation operations, making them less practical for T2V generation.
  \vspace{-0.5cm}
\paragraph{Direct optimization for guided diffusion models.} Recent studies have demonstrated that optimizing through diffusion sampling is feasible with proper GPU memory management. Direct Optimization of Diffusion Latents (DOODL)~\cite{wallace2023doodl} builds on the EDICT method~\cite{wallace2022edict}, using coupling layers to create a fully invertible sampling process. DOODL backpropagates through EDICT to refine initial noise latents, enhancing high-level features such as CLIP guidance and aesthetic quality in images. However, unlike TITAN-Guide, DOODL has several limitations due to its dependence on EDICT, including: 1) dependence on invertible sampling algorithms and, 2) twice the model sizes for forward and reverse sampling. These two limitations lead to increased latency and memory consumption. Another method, DITTO~\cite{novack2024ditto}, employs a similar optimization-based guidance strategy but faces memory consumption challenges, much like DOODL. Although DITTO shares similar objectives with DOODL, its applicability is limited to sound data with  smaller dimension compared with the visual domain and with a small number of denoising steps (e.g., 10–12 steps). Both methods struggle to meet the memory constraints required for generating high-resolution, multi-frame videos on a single consumer GPU \ie $\sim$ 24GB.

\begin{figure*}[!t]
    \centering
    \includegraphics[width=1.0\textwidth]{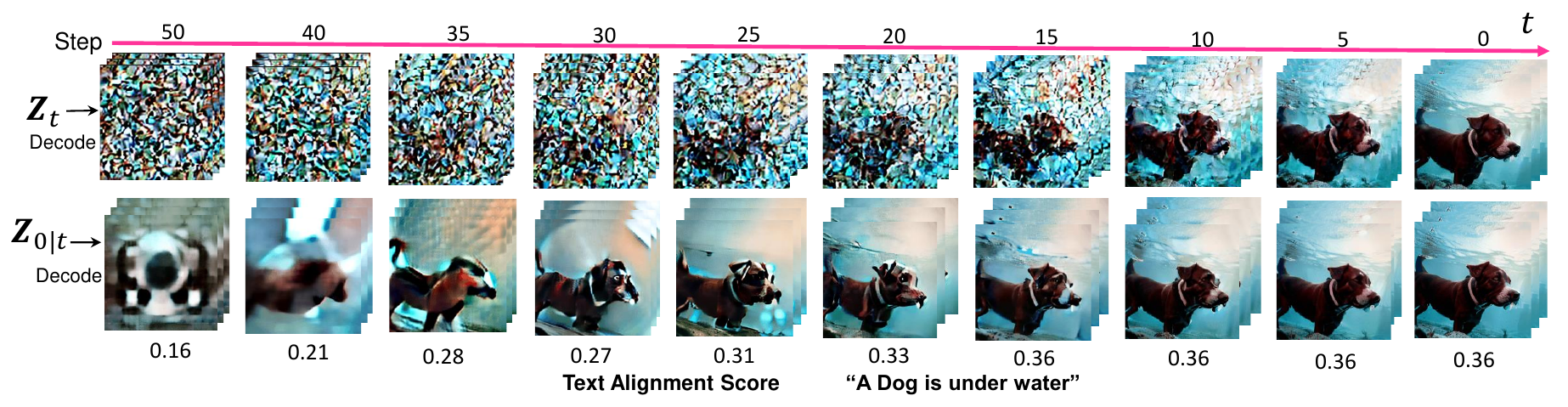}
    \caption{
Projection of clean data estimation \(\Vec{z}_{0|t}\) and latent noise \(\Vec{z}_{t}\) onto video \(\Mat{X}\). Since \textit{off-the-shelf} classifier models are not specifically designed for corrupted visual data, evaluating  clean estimation \(\Vec{z}_{0|t}\) in early iterations propagates false signals, as indicated by low text alignment scores (\eg, using ImageBind~\cite{girdhar2023imagebind}). However, steering diffusion latents in later iterations has less impact than in earlier ones, as video's appearance has already taken shape.
    }
    \label{fig:step_problem}
\end{figure*}

\paragraph{Forward gradient descents.} Though reverse mode \ie, backpropagation is proven effective to update neural network models, this operation is memory extensive and non-efficient for optimization. Most guidance diffusion methods apply reverse mode Automatic Differentiation (AD) involving data-flow reversal which is costly.  Forward mode is used to estimate gradients for optimization. A notable method called forward gradient descents~\cite{baydin2022fwdgrad} provides an alternative on computing gradients via forward AD. The forward gradient method offers a training approach that aligns more closely with biological learning processes compared with backpropagation. 



\section{Preliminaries}

\paragraph{Diffusion models.} Our pipeline incorporates denoising diffusion models, with a particular focus on T2V models. We operate in the latent space, utilizing an encoder $E_\phi$ and a decoder $D_\phi$ to process video inputs by encoding them into a compressed latent representation $\Vec{Z}_t \in \mathbb{R}^{{F} \times {C} \times {H} \times {W}}$ and decode the latent to the original space  $\Vec{X} \in \mathbb{R}^{F^\prime\times C^\prime \times H^\prime \times W^\prime}$.
Noisy data are obtained by adding Gaussian noise as follows:
\begin{equation}
\Vec{Z}_t = \sqrt{\bar\alpha_t}\Vec{Z}_0 + \sqrt{1 - \bar\alpha_t} \epsilon,
\end{equation}
where $\epsilon \sim \mathcal{N}(0, \mathbf{I})$ and $\bar\alpha_t$ is the scaling factor at time $t$. The denoising process is a backward diffusion employing a denoiser $\eps_\theta$ to obtain $\Vec{Z}_{t-1}$ as follows:
\begin{equation}
    p_\theta (\Vec{Z}_{t-1} | \Vec{Z}_t) = \mathcal{N}(\Vec{Z}_{t-1};\Vec\mu_\theta(\Vec{Z}_t, t, \Vec{c}), \Vec\Sigma_{\theta}(\Vec{Z}_t, t, \Vec{c})),
\end{equation}
where $\Vec{c}$ is a conditional embedding (\eg, text), and $\Vec\mu_\theta$ and $\Vec\Sigma_\theta$ are the mean and variance determined by a denoiser $\epsilon_\theta(\Vec{Z}_t, t, \Vec{c})$, which is parameterized by $\theta$.
\paragraph{Classifier guidance.} Off-the-shelf diffusion models have limitations in expressing the generated results on a certain direction. Therefore, classifier guidance was proposed to steer the output of a diffusion model. Several recent studies on classifier guidance~\cite{song2023ugd,yu2023freedom,ye2024tfg} aim to leverage pretrained diffusion models to control conditional generation tasks. This concept is  derived from the conditional score function:
\begin{equation}
    \nabla_{\Vec{Z}_{t}} \log p(\Vec{Z}_t | \Vec{y}) = \nabla_{\Vec{Z}_{t}} \log p(\Vec{Z}_t) + \nabla_{\Vec{Z}_{t}} \mathcal{L}(\Vec{Z}_t, \Vec{y}, \psi),
\end{equation}
where $\mathcal{L}$ is an objective function, $\psi$ is  a classifier model, and $\Vec{y}$ is a target for alignment (\eg, text, style, image, audio). However, the pretrained guidance loss function is usually defined solely on clean data $\Vec{Z}=0$ rather than on noisy data $\Vec{Z}_t$. We typically only have access to a pretrained classifier model $\psi$ trained on clean data, instead of one trained on noisy data. 
With some recent classifier guidance methods~\cite{song2023ugd,yu2023freedom,ye2024tfg}  assume that the clean data estimation is accessible via a time-dependent differentiable function.  The update rule for the denoise samples can be expressed as:
\begin{equation}
    \Vec{Z}_t \xleftarrow{} \Vec{Z}_t - \lambda_t \nabla_{\Vec{Z}_{t}} \mathcal{L}(\Vec{Z}_{0|t},\Vec{y},\psi),
\end{equation}
where  $\Vec{Z}_{0|t}$ can be estimated from the Tweedie's formula~\cite{Efron2011TweediesFA} as follows:
\begin{equation}
    \Vec{Z}_{0|t} = \frac{1}{\sqrt{\bar\alpha_t}}\big(\Vec{Z}_t - \sqrt{1-\bar\alpha_t}\epsilon_\theta(\Vec{Z}_t, t, \Vec{c})\big).
\end{equation}
While the clean data estimation $\Vec{Z}_{0|t}$ offers a promising method for obtaining gradient signals compared with noisy data $\Vec{Z}_t$, this remains problematic during the early sampling stages, as illustrated in Fig.~\ref{fig:step_problem}. Visually, the estimated clean data $\Vec{Z}_{0|t}$, when projected into the input space through $D_\phi$, lacks sufficient update signals for effective  guidance.

\begin{figure*}[!t]
    \centering
    \includegraphics[width=0.98\textwidth]{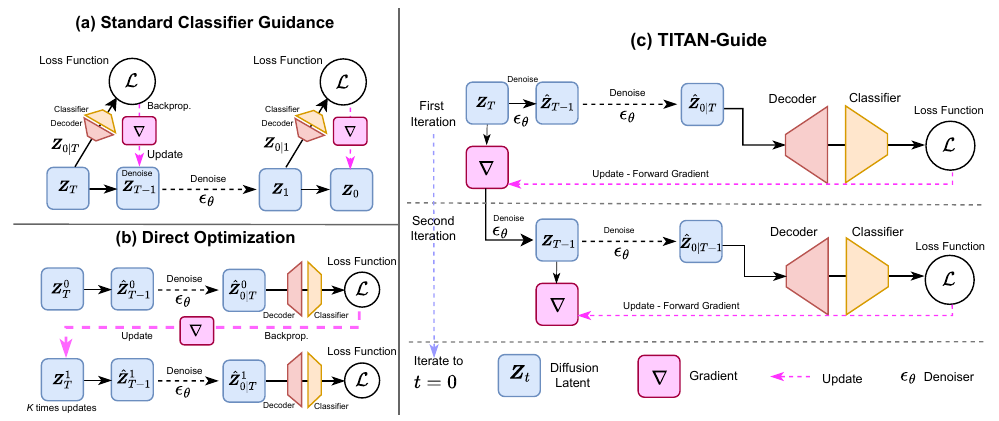}
    \caption{ Comparison of TITAN-Guide with previous diffusion guidance methods. (a) Standard classifier guidance, with which latents are updated at each iteration step~\cite{ye2024tfg,he2024mpgd,song2020classifierguidance,yu2023freedom}. (b) Direct optimization approach~\cite{wallace2023doodl}, where latents are updated only after completing all iteration steps. (c) Illustration of TITAN-Guide. Latent input~$\Vec{Z}_t$ is updated by unrolling diffusion process to $t=0$ though a denoiser $\Vec{\epsilon}_\theta$. Update is determined from gradient with respect to $\Vec{Z}_t$, based on the loss function $\mathcal{L}$. TITAN-Guide eliminates the need for backpropagation by leveraging forward gradients, enabling efficient memory usage.   }
    \label{fig:ourframework}
\end{figure*}

\paragraph{Forward gradients.} For simplicity, we use a general form of the forward gradient applied to the input with $d$-dimension, while this notation can be easily extended to the latent of a video.  
Formally, we define the forward gradient method as $G: \mathbb{R}^d \rightarrow \mathbb{R}^d$.  
An efficient and accurate $G(\Vec{Z}) \in \mathbb{R}^d$ of a gradient $\nabla {f}(\Vec{Z}) \in \mathbb{R}^d$, where the objective function is defined as ${f} : \mathbb{R}^d \rightarrow \mathbb{R}$. To approximate the gradient, we formally have:
\begin{equation}
    G(\Vec{Z}) = \langle \nabla  {f}(\Vec{Z}), \Vec{V}\rangle \Vec{V} \approx \nabla  {f}(\Vec{Z}),
\end{equation}
where $\Vec{V}$ is a gradient guess vector that projects the gradient target, also called as tangent.
 In forward mode AD, we can provide a directional derivative without having to explicitly compute $\nabla{f}(\Vec{Z})$. This approach induces lower computational overhead compared with the standard forward process. This routine operates in a single forward pass, eliminating the need for a backward pass and waiting time for an update signal, in contrast to backpropagation. 
Previous guided diffusion models (\eg, DOODL~\cite{wallace2023doodl}) still require backpropagation, while TITAN-Guide does not need to execute backward pass in computing the gradients. 



\section{Proposed Method: TITAN-Guide}
As described previously, previous methods for guided diffusion models~\cite{wallace2023doodl,yu2023freedom,ye2024tfg,he2024mpgd} depend heavily on the backward process when computing the gradients to update the noise latent. Moreover, this inefficient process is not directly applicable when guiding memory-intensive diffusion models \eg, T2V models. To overcome this challenge, TITAN-Guide improves classifier-based diffusion guidance. In particular, it guides T2V diffusion models via forward gradient (see  Fig.~\ref{fig:ourframework} for a conceptual diagram).

\subsection{Guided Text-to-Video Diffusion Models}
We aim to process the video latent representation $\Vec{Z}$ with a temporal frame ${F}$. To achieve diffusion guidance, the latent noise must first be projected back into the original space $\Mat{X}$ using a decoder $D_{\phi}$. After reconstruction, the video is then passed through a classifier model to obtain update direction for the latent representation.

\paragraph{On manipulating latents.} 
In classifier guidance, the noise latent $\Vec{Z}_t$ is updated during the denoising steps with an objective function $\mathcal{L}(\Vec{Z}_{0|t}, \Vec{y}, \psi)$. 
We aim to investigate latent diffusion equipped with a $D_\phi$, by defining an objective function $\mathnormal{f}$. 
Despite the success in image domain, the decoders in T2V diffusion models~\cite{yang2024cogvideox,guo2023animatediff} are quite costly in terms of GPU memory usage, especially to project a latent to the original space with ${F}$ number of frames.  

As mentioned previously, clean data estimation as performed in previous studies~\cite{ye2024tfg,yu2023freedom,he2024mpgd} propagate inaccurate signals for latent updates. A more favorable approach to this problem is  sampling until iteration $t=0$ as follows:
\begin{equation}
    \hat{\Vec{Z}}_{0|t} = \texttt{Sample}_{t \xrightarrow{} 0}(\epsilon_\theta, \Vec{Z}_t, t, \Vec{c}).
\end{equation}
In the standard guided diffusion process, iterating diffusion steps to $t=0$ is only feasible for a limited range as in text-to-image diffusion models with a small subset of sampling (\eg, 12 steps) as in previous studies~\cite{wallace2023doodl,novack2024ditto}. 
Finally, we update to the latent diffusion as follows:
\begin{equation}
  \Vec{Z}_{t} \leftarrow  {\Vec{Z}}_{t} - \lambda_t \nabla_{\Vec{Z}_t} f\big((D_\phi(\hat{\Vec{Z}}_{0|t}),  \Vec{y}, \psi \big),
\end{equation}
where $\lambda_t$ is the learning rate, and in our implementation, $\psi$ is a classifier model \eg, ImageBind~\cite{girdhar2023imagebind}, DOVER~\cite{wu2023dover}, or Style-CLIP~\cite{patashnik2021styleclip}.  

\paragraph{Forward gradient descents.} 
Previous classifier-guidance methods traversing to $t=0$ (\eg, DOODL~\cite{wallace2023doodl}) cannot afford for expensive computation of the decoding process in T2V generation. 
Since both the decoding process and sampling from $t$ to $0$ are memory-intensive, we investigate a new classifier guidance technique that leverages gradient approximation methods, such as forward gradient~\cite{baydin2022fwdgrad,ren2022scaling,fournier2023can}, to update the latent diffusion. This technique is more efficient than computing gradients via backpropagation, enabling direct sampling the clean latent output $\hat{\Vec{Z}}_{0|t}$ instead of using the estimated clean output $\Vec{Z}_{0|t}$. With forward gradient descents, we estimate the gradient at time $t$ as $
    \Vec{G}_{t} =  \langle \nabla f\big((D_\phi(\hat{\Vec{Z}}_{0|t}),  \Vec{y}, \psi \big), \Vec{V}_{t}\rangle \Vec{V}_{t}$. Algorithm~\ref{code:algorithm1} outlines the detailed steps of TITAN-Guide. 

\begin{algorithm}[!t]
\caption{TITAN-Guide}
{\bf Input:} denoiser $\epsilon_\theta$, latent input $\Vec{Z}_t$, learning rate $\lambda_t$, num. of iter. $T$, condition for diffusion $\Vec{c}$, target $\Vec{y}$, classifier $\psi$, decoder $D_\phi$ 
\begin{algorithmic}[1]
\State // Traverse to $\Vec{Z}_0$ and pass to the classifier model
\Function{$f$}{$ \epsilon_\theta, \Vec{Z}_t, t, \Vec{c}, \Vec{y}, \psi, D_\phi$}:
    \State $\hat{\Vec{Z}}_{0|t} \leftarrow \texttt{Sample}_{t \xrightarrow{} 0}(\epsilon_\theta, \Vec{Z}_t, t, \Vec{c})$\
    \State $\Vec{\bar{y}} \leftarrow \psi\big(D_\phi(\hat{\Vec{Z}}_{0|t})$\big)\
    \State Compute loss between $\Vec{\bar{y}}$ and $\Vec{y}$
\EndFunction \State \textbf{end function}
\State // Denoising steps with guidance and forward gradient
\State $\Vec{Z}_T \sim \mathcal{N}(0, \textbf{I})$
\For{$t=T \textrm{ to } 1 $} 
    \State $\Vec{Z}_{t}  \leftarrow \texttt{Sample}_{t+1 \xrightarrow{} t}(\epsilon_\theta, \Vec{Z}_{t+1}, t, \Vec{c})$\
    \State  Initialize $\Vec{V}_t$ 
    \State $\texttt{Params}=(\epsilon_\theta, \Vec{Z}_t, t, \Vec{c}, \Vec{y}, \psi, D_\phi)$
    \State $\mathcal{L},\; h_t \leftarrow f(\texttt{Params}),\; \langle \nabla{f}(\texttt{Params}), \Vec{V}_{t}\rangle$\
    \State $\Vec{G}_{t} \leftarrow  h_t \Vec{V}_t $\ 
    \State $\Vec{Z}_{t} \leftarrow \Vec{Z}_{t} - \lambda_{t} \Vec{G}_t$
\EndFor \State \textbf{end for}

\end{algorithmic}
\label{code:algorithm1}
\end{algorithm}


\subsection{The Many Faces of Gradient Guesses} 
Regarding the selection of direction distribution, the gradient guess 
$\Vec{V}$ in forward gradient exhibits a certain degree of randomness, allowing for variability in its estimation. In fact, the approximation of this guess can be determined through either probabilistic methods, which incorporate stochastic elements, or deterministic methods, which follow a fixed, directed computation. Each method offers distinct implications depending on the context in which the gradient estimation is applied.
\paragraph{Random guesses.} In the previous study~\cite{baydin2022fwdgrad}, the forward gradient method perturbed samples by drawing $\Vec{V}_t$ from a zero-mean, unit covariance probability distribution. To better estimate gradients for classifier-diffusion guidance, we specifically propose using isotropic Gaussians $\Vec{V}_t \sim \mathcal{N}(0, \mathbf{I})$. In both cases, $\Vec{G}$ is an unbiased estimate of $\nabla f$. However, such estimates potentially have
high variance, leading to errors in individual gradient
estimates.

\paragraph{Score-based guesses.} In the diffusion process, a clean distribution can be obtained by iteratively applying the time-dependent score function, which is parameterized using a denoiser $\epsilon_\theta(\Vec{Z}_t, t, \Vec{c})$.  Essentially, the denoiser estimates Gaussian noise for each noisy latent $\Vec{Z}_t$. To improve the quality of the gradient for latent updates, we incorporate deterministic gradient guesses from the denoiser. In the shed-of-light estimated Gaussian noise, we propose score-based guesses that leverages the output of a denoiser, providing a direct estimate of denoising steps. Before we perform forward gradient descents, the direction is estimated through deterministic guesses by having the projection direction:
\begin{equation}
    \Vec{V}_t = \frac{\epsilon_\theta(\Vec{Z}_t, t, \Vec{c})}{\|  \epsilon_\theta(\Vec{Z}_t, t, \Vec{c}) \| }.
\end{equation}
The key idea is that the denoiser's output is conditioned on the input latent and the time $t$ at each diffusion iteration, resulting in a more deterministic approximation than uncorrelated random noise, which has high variance. 

\paragraph{Sampled gradient guesses.} As we can initially estimate clean outputs using the Tweedie's formula~\cite{Efron2011TweediesFA}, estimating gradients using a subset of frames is feasible. Let us denote $\Vec{Z}^{\mathcal{F}}_{0|t}$ as a small subset of latents corresponding to the timeframe, where $\mathcal{F} \ll {F}$.   Thus, we can define $\Mat{V}_t$ by sampling from $\nabla_{\Vec{Z}^{\mathcal{F}}_{0|t} } f\big((D_\phi({\Vec{Z}}^{\mathcal{F}}_{0|t}), \Vec{y}, \psi \big)$. We normalize the sampled gradient. Then, we expand the gradient sampling estimation to match the number of ${F}$ frames. This approach is more lightweight compared with computing the gradient with a full number of frames.





\begin{figure*}[t]
    \centering
    \includegraphics[width=0.999\textwidth]{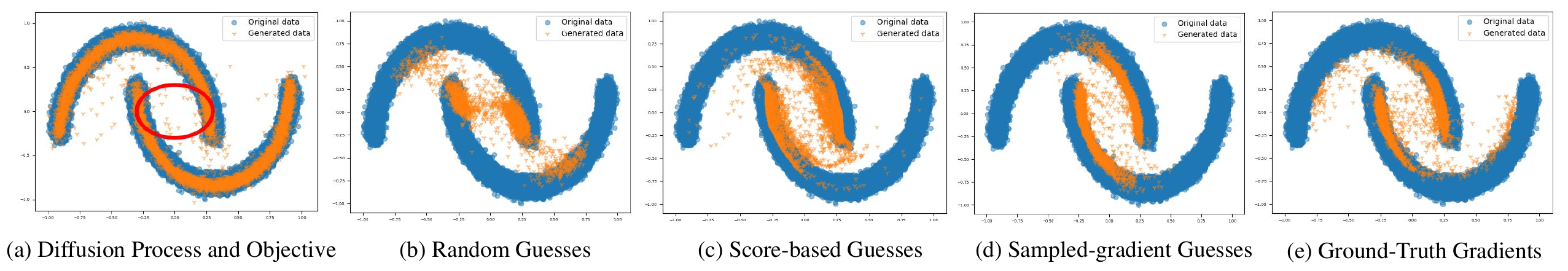}
    \vspace{-0.5cm}
\caption{Experiments on toy data (\ie, Moons) using diffusion models, with original data in \textcolor{cornflowerblue}{blue} and generated samples in \textcolor{orange}{orange}. We compare guided diffusion against true gradients: (a) Original distribution and generated samples, with  objective function in \textcolor{red}{red}. TITAN-Guide with random guesses (b), score-based guesses (c), and sampled-gradient guesses (d). (e) True gradients of guided diffusion.}
\vspace{-0.12cm}
     \label{fig:toy_data}    
\end{figure*}

\section{Experiments}
We conducted a comprehensive evaluation of TITAN-Guide for T2V diffusion models across multiple benchmarks. In addition to the performance comparison, we also conducted a toy experiment, and various conditional settings. The generated samples are available at our project page\footnote{\url{https://titanguide.github.io/}}.

\subsection{Implementation Details}
\paragraph{Datasets.} 
For the datasets, we used VGG-Sound~\cite{chen20vggsound} and TFG-1000 prompts~\cite{ye2024tfg}. For the VGG-Sound benchmark, we followed the experimental setup in~\cite{xing24seeing} to ensure consistency and comparability. However, since the original set of captions and reference videos used in~\cite{xing24seeing} has not been publicly released, we constructed an evaluation dataset by randomly selecting 3,000 video-audio pairs from  VGG-Sound for evaluation. Each selected instance is then assigned a corresponding caption generated using automated captioning models~\cite{Qwen-Audio,Qwen-VL}. 
To analyze the performance of TITAN-Guide, we conducted the toy experiment using the Moons dataset to simulate classifier guidance in diffusion, applied for various gradient guesses and exact gradients.

\begin{table*}[!t]
 \centering
    \resizebox{1\textwidth}{!}{
    \Large\addtolength{\tabcolsep}{-0.1pt}
\begin{tabular}{lcccccccccccccc}
\toprule
\multirow{2}{*}{Method}                &\multirow{2}{*}{Memory $\downarrow$ } & \multicolumn{7}{c}{Audio-Video Align.}   & \multicolumn{2}{c}{Style Guidance}  & \multicolumn{2}{c}{Aesthetic Guidance} & \multicolumn{2}{c}{Frame Interpolation}  
\\ 
& & FVD$\downarrow$    & IB – AV$\uparrow$  & IB – TV$\uparrow$ & AV-Align$\uparrow$ & KVD$\downarrow$   & F. Cons.$\uparrow$ & DOVER$\uparrow$ & ED$\downarrow$ & IB-TV$\uparrow$ &DOVER$\uparrow$ &IB-TV$\uparrow$  & FVD$\downarrow$ & F. Cons.$\uparrow$ \\
\hline
Seeing-and-Hearing~\cite{xing24seeing}         &59.6GB                   & 392.23& {0.218}                                                                    & {0.315}                                                                    & 0.470     & 30.01& 0.953                          & 0.433     & - & - & - & - & - & -                                                                        \\
FreeDoM~\cite{yu2023freedom}            &60.2GB                 & 390.21& {0.218}                                                                    & {0.314}                                                                    & 0.474     & 32.15& 0.955                         & 0.455    & 9.95 & 0.21 & 0.60 & 0.314 & 170.77 & 0.934                                                                         \\

MPGD~\cite{he2024mpgd}                &41.0GB           & 366.02& {0.217}                                                                    & {0.315}                                                                    & 0.472     & 28.91& 0.958                         & 0.457       
 & 9.33 & 0.24 & 0.58 & 0.318 & 169.88 & 0.932                
\\
TFG~\cite{ye2024tfg}              &40.9GB              & 365.19& {0.217}                                                                    & {0.315}                                                                    & 0.474     & 30.42 & 0.959                          & 0.457 
& \textbf{8.92} & 0.28 & 0.52 & 0.319 & 170.89 & 0.916
\\
DOODL~\cite{wallace2023doodl}       &67.2GB                     & 334.21& {0.218}                                                                    & {0.315}                                                                    & 0.477     & 27.10 & 0.960                          & 0.453  
& 9.83 & 0.27 & 0.62 & 0.318 & 169.21 & 0.934\\
 \rowcolor{aliceblue} TITAN-Guide (ours) & & & & & & & & & & & & & &\\ 
 \rowcolor{aliceblue} \; - Random guesses      &\textbf{20.6GB}              & 348.26 & \textbf{0.218}                                                                    & \textbf{0.315}                                                                    & 0.473     & 28.93& {0.954}                          & 0.445  & 9.98 & 0.29 & 0.55 & 0.319 & 171.42 & 0.920                                                                            \\
 \rowcolor{aliceblue} \; - Score-based guesses &\textbf{20.6GB} & {333.42} & 0.215                                                                    & 0.312                                                                    & \textbf{0.480}     & {26.95}& 0.955                          & {0.461}  
 & 9.66 & 0.28 & 0.52 & 0.316 & 170.11 & 0.930 \\  
  \rowcolor{aliceblue} \; - Sampled gradient guesses &\textbf{20.6GB} & \textbf{331.11} & 0.216                                                                   & 0.314                                                                    & \textbf{0.480}     & \textbf{26.19}& \textbf{0.961}                          & \textbf{0.465} 
   & 9.16 & \textbf{0.29} & \textbf{0.65} & \textbf{0.320} & \textbf{168.88} & \textbf{0.934}\\  
 \bottomrule
 \bottomrule
\end{tabular}
}
\caption{Evaluation results across various tasks (\eg, audio-video alignment, style guidance, aesthetic guidance and frame interpolation) and various metrics for assessing the quality of 256x256 generated videos. 
}
\label{table:vggsound_result}
\end{table*}

\begin{figure*}[h]
    \centering
    \includegraphics[width=1.01\textwidth]{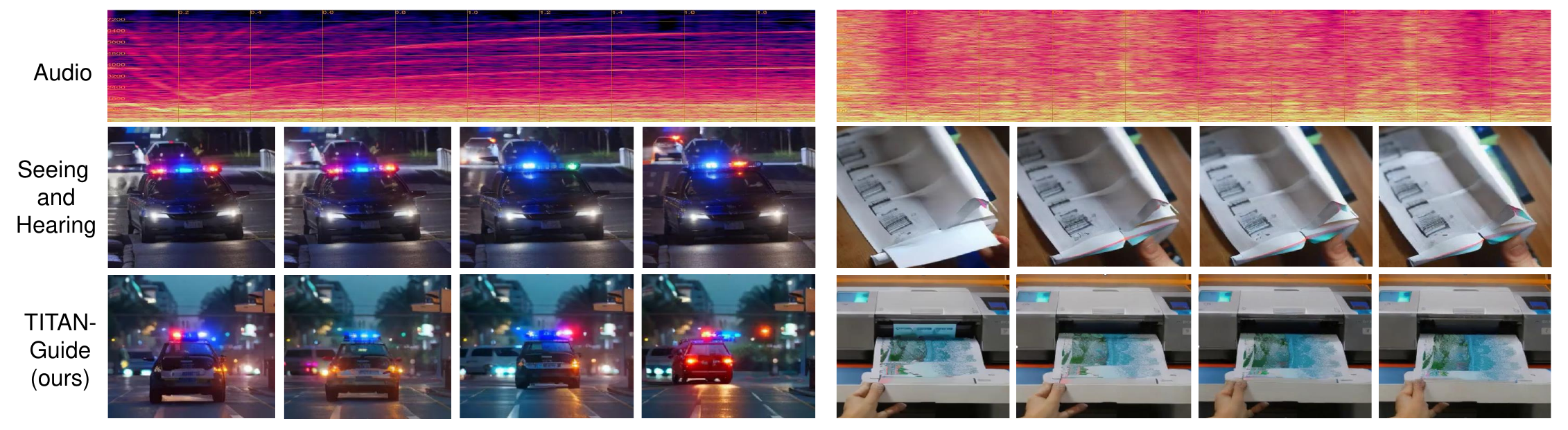}
    \caption{
Results of our proposed approach compared with Seeing-and-Hearing~\cite{xing24seeing} on audio-to-video tasks. TITAN-Guide consistently generated videos that aligned well with  reference audio. Compared with the baseline method, TITAN-Guide not only maintains audio alignment but also captures finer visual nuances and object actions more effectively. For example, in scene with  police siren and  moving car engine, our approach accurately depicted the car in motion, whereas Seeing-and-Hearing failed to show any movement. }
    \label{fig:results_audio_video}
\end{figure*} 

\paragraph{Settings.} For evaluation, we used  AnimateDiff~\cite{guo2023animatediff} and CogVideoX~\cite{yang2024cogvideox} as the base T2V diffusion model, which are noth effective in performance and GPU memory consumption. This was to make comparison with the guided-diffusion baselines feasible. 
For classifier guidance, we employed off-the-shelf discriminative models: 1) ImageBind~\cite{girdhar2023imagebind} for audio-text-video alignment tasks, 2) DOVER~\cite{wu2023dover} to enhance aesthetic aspects \eg, semantics and composition, and 3) Style-CLIP~\cite{patashnik2021styleclip} to control the style of generated videos.
We conducted our evaluations at a resolution of 256×256 with 16 frames for 2s using AnimateDiff and 256×384 with 32 frames for 2s  using CogVideoX. The baseline methods are unable to operate beyond 256×256 resolution on consumer-grade GPUs due to memory constraints. In contrast, TITAN-Guide supports resolutions up to 384×384 while remaining feasible. For diffusion-based generation, we set AnimateDiff to execute 20 denoising iterations. Please refer to our supplementary material for more details.
\paragraph{Guidance tasks.} 
We used three guidance  tasks: \textbf{aesthetic score guidance}, \textbf{style guidance}, and \textbf{audio-video alignment} with DOVER~\cite{wu2023dover}, Style-CLIP~\cite{patashnik2021styleclip}, and ImageBind~\cite{girdhar2023imagebind} as the guidance models, respectively. 
Additionally, \textbf{frame interpolation} was also performed to generate intermediate frames. We used TFG-1000Prompt~\cite{ye2024tfg} for aesthetic-score and style guidance, and VGG-Sound~\cite{chen20vggsound} for multi-modal alignment and frame interpolation. 

\paragraph{Baselines.} 
In our experiments, we conducted a comprehensive comparison of our proposed method TITAN-Guide against DOODL~\cite{wallace2023doodl}, TFG~\cite{ye2024tfg}, MPGD~\cite{he2024mpgd}, and FreeDoM~\cite{yu2023freedom} for T2V generation, specifically evaluating video quality scores. These methods were chosen as they represent distinct methods within the classifier-guidance framework, providing a well-rounded basis for comparison. We extended our evaluation to video generation with audio-video alignment by benchmarking against Seeing and Hearing~\cite{xing24seeing}, which was designed to enhance multimodal synchronization. To ensure a fair and consistent comparison, we implemented   all baselines under our experimental settings and environment, maintaining identical configurations, including the same seed number, across all runs.

\paragraph{Metrics.}
We evaluated video quality using Fréchet Video Distance (FVD) and Kernel Video Distance (KVD)~\cite{unterthiner2018fvd}, where lower scores indicate better realism and coherence. Aesthetic quality was assessed with DOVER scores~\cite{wu2023dover}, ensuring visually appealing and well-composed outputs. For multi-modal alignment, we used ImageBind scores~\cite{girdhar2023imagebind} to measure text-video (IB-TV) and audio-video (IB-AV) correspondences. Frame consistency was analyzed by comparing image representations across consecutive frames~\cite{wu2023consistency,wu2023tune}, while AV-Align scores~\cite{yariv2023tempo} were used to evaluate audio-video synchronization. These metrics provide a comprehensive assessment of realism, aesthetics, alignment, and temporal consistency. Also, we used Euclidean Distance (ED) in the feature space to measure similarity in style guidance.

\subsection{Results}
\paragraph{Toy experiment.} We analyzed guided diffusion models with a toy experiment on the Moons dataset in Fig.~\ref{fig:toy_data}, using a circular objective function, $x^2 + y^2 - 0.3$. TITAN-Guide, using both random and score-based guesses, effectively approximated the objective function similar to ground truth gradients. While differing from the exact gradient, guidance remains feasible. Additionally, we observed that random guesses tend to concentrate outputs within a smaller region compared with score-based guesses.

\paragraph{Results on multi-modal alignment.} Table~\ref{table:vggsound_result} presents a comparison of TITAN-Guide with baselines using AnimateDiff on the VGG-Sound dataset~\cite{chen20vggsound}. TITAN-Guide, when using random guesses, improved the FVD score by around 40 points, while with score-based guesses, it achieved an FVD score of 333.42, demonstrating its effectiveness in generating high-quality videos. Using the sampled gradient guesses, we could achieve the lowest FVD of 331.11. Additionally, we observed a notable improvement in the AV-Align score compared with the baseline methods. We attributed this improvement to our iterative approach, which traverses and refines the output toward $t=0$ to obtain a clean data distribution. In contrast, the baseline method relied on an estimated clean output $\Vec{Z}_{0|t}$ which has non-proper visual appearance in early iterations. 
Fig.~\ref{fig:results_audio_video} shows a qualitative comparison with Seeing and Hearing~\cite{xing24seeing}. We observed similar behaviours with CogVideoX as a T2V backbone model as shown in Table~\ref{table:cogvid2}.  


\begin{figure}[t]
    \centering
    \includegraphics[width=.492\textwidth]{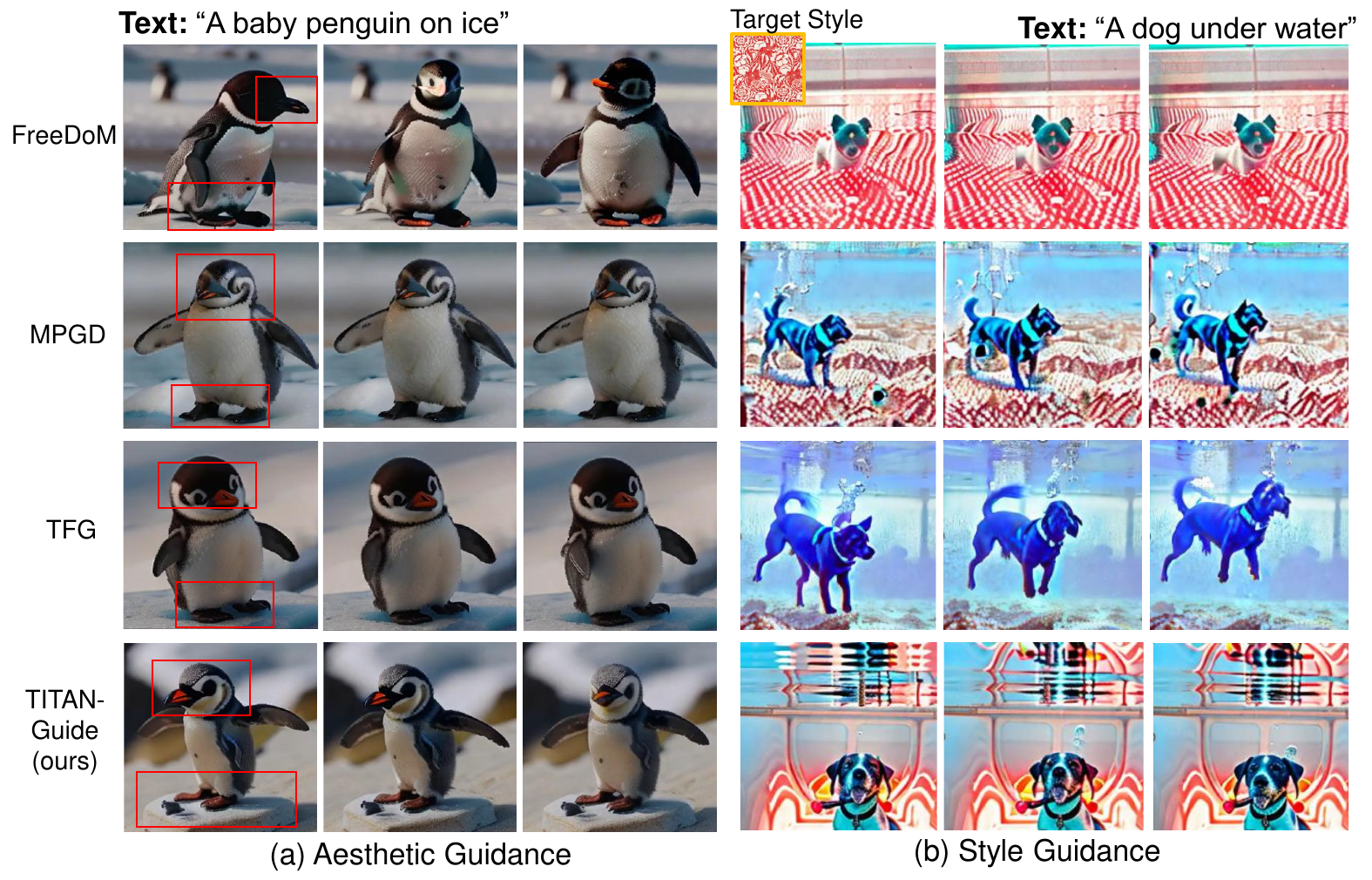}
    \caption{Qualitative comparisons on T2V guidance: (a) Aesthetic alignment using DOVER scores (e.g., semantics, composition). (b) Style alignment. See our supplementary material for videos.}
    \label{fig:style_aesthetic_result}
\end{figure}

\paragraph{Results on aesthetic-score guidance.} Fig.~\ref{fig:style_aesthetic_result}  compares TITAN-Guide with FreeDoM, MPGD, and TFG, highlighting key differences in frame consistency, compositional quality, and object alignment.
FreeDoM struggled with temporal consistency, leading to different faces between frames. MPGD produced videos with poor compositionality, affecting spatial arrangement in the head. TFG misaligned objects components \eg, a penguin’s eyes and feet.
In contrast, TITAN-Guide generated well-composed videos with consistent frames and accurate object alignment, demonstrating its effectiveness in maintaining both spatial and temporal coherence. In Table~\ref{table:vggsound_result}, TITAN-Guide performed better with the highest DOVER score.

\paragraph{Results on style guidance.} Our analysis highlights key differences in style transfer and object preservation across the baselines. While FreeDoM~\cite{yu2023freedom} is capable of reflecting the style of the target image to some extent, it suffered from a loss of text alignment, leading to noticeable degradation in key visual elements \eg, water and dog details. MPGD~\cite{he2024mpgd} managed to transfer the style to a limited degree but at the cost of deteriorating the overall object composition. Similarly, TFG~\cite{ye2024tfg} fails to maintain the integrity of object structures, resulting in inconsistencies in the generated content. TITAN-Guide demonstrated its effectiveness, achieving a balanced outcome where both object composition and style transfer are well-preserved as shown in Fig.~\ref{fig:style_aesthetic_result}. 

\paragraph{Results on frame interpolation.} Additionally, we present the results on frame interpolation. As shown in Table~\ref{table:vggsound_result},TITAN-Guide consistently surpassed previous approaches in performance. Notably, TITAN-Guide achieved an FVD score of 168.88, marking the best result.

\begin{table}[t]
 \centering
    \resizebox{0.48\textwidth}{!}{
    \Large\addtolength{\tabcolsep}{-0.3pt}
\begin{tabular}{lcccccc}
\toprule
\multirow{2}{*}{Method}     &\multicolumn{2}{c}{Style Guidance} &\multicolumn{2}{c}{Aesthetic Guidance} &\multicolumn{2}{c}{Audio-Video Align.}  
 \\    
 &ED$\downarrow$ &IB – TV$\uparrow$ &DOVER$\uparrow$ &IB – TV$\uparrow$   & FVD$\downarrow$ &AV-Align$\uparrow$ \\
\hline
TFG~\cite{ye2024tfg}
           &   9.33                                                               &  0.291     & 0.513   &  \textbf{0.335}                      & 343.28 
&0.421 
 \\
 \rowcolor{aliceblue} TITAN-Guide (ours)  & & & & & &\\
 \rowcolor{aliceblue}  - Random guesses      &   10.47                                                                  &  \textbf{0.335}   &  0.508  &  0.315                         & 342.36    &\textbf{0.433}                                                    \\
 \rowcolor{aliceblue}   - Score-based guesses & 10.59  & 0.333  & 0.506  & 0.334                          & 
328.96   &0.425   \\  
  \rowcolor{aliceblue} - Sampled gradient guesses & \textbf{7.93}   &  0.300 &   \textbf{0.568}                                                                & \textbf{0.335}                             & \textbf{318.99}   &0.427   \\  
 \bottomrule
\end{tabular}
}
\caption{Evaluation results using {CogVideoX} T2V on style, aesthetic, and audio-video guidance tasks on 32 frames.}  
\vspace{-0.15cm}
\label{table:cogvid2}
\end{table}

\paragraph{Ablation on number of iterations}
We conducted an experiment by varying the number of iterations using a random guessing strategy (see Table~\ref{table:numiter}). The results indicate that increasing the number of iterations leads to only marginal improvements in performance, suggesting that the random guessing strategy quickly reaches a marginal performance gain from additional computation.

\begin{table}[t]
 \centering
    \resizebox{0.49\textwidth}{!}{
    \Large\addtolength{\tabcolsep}{.3pt}
\begin{tabular}{lcccccc}
\toprule
\multirow{2}{*}{Method}     &\multicolumn{6}{c}{Audio-Video Alignment}  \\    
 &FVD$\downarrow$ &KVD$\downarrow$ &IB-AV$\uparrow$ &IB-TV$\uparrow$ &AV-Align$\uparrow$ &F.Cons.$\uparrow$   \\
\hline
    20 iterations & 348.26 & 28.93                                                              & 0.218 & 0.315   &  0.473 &  0.954                      \\  
  50 iterations & 340.21 & 28.67                                                             & 0.219 & 0.318    &  0.476 &   0.954   \\
 100 iterations & 350.52 & 30.42  & 0.219 & 0.318   &  0.476 &   0.955  \\
 \bottomrule 
\end{tabular}
}
\caption{Ablation for TITAN-Guide with various numbers of iterations on audio-video alignment task. }
\vspace{-0.15cm}
\label{table:numiter}
\end{table}

\begin{table}[t]
 \centering
    \resizebox{0.48\textwidth}{!}{
    \Large\addtolength{\tabcolsep}{0.5pt}
\begin{tabular}{lcccccc}
\toprule
 Method   & FVD$\downarrow$ & KVD$\downarrow$ 
& AV-Align$\uparrow$    
& DOVER$\uparrow$ & IB-AV$\uparrow$ & IB-TV $\uparrow$ \\
\hline
Random guesses & 320.05 & 27.14                                                                    & 0.475                                                                   
& 0.675      & 0.217 & 0.316                                                                \\
Score-based guesses & 319.01 & 26.93                                                 & 0.481 & 0.677     & 0.216 & 0.317                  
                                                                   \\
Sampled gradient guesses & \textbf{317.05} & \textbf{26.91}                                                 &\textbf{0.481}  &\textbf{0.677}  & \textbf{0.219} & \textbf{0.317}                 
                                                                      \\
\bottomrule
\end{tabular}
}
\vspace{-0.1cm}
\caption{Evaluation results on a higher resolution   (384$\times$384) for audio-video alignment task with various types of guesses.}
\label{table:vgg_highres}
\end{table}

\begin{figure}[t]
    \centering
 \subfloat[256x256 Resolution]{
        \includegraphics[width=0.222\textwidth]{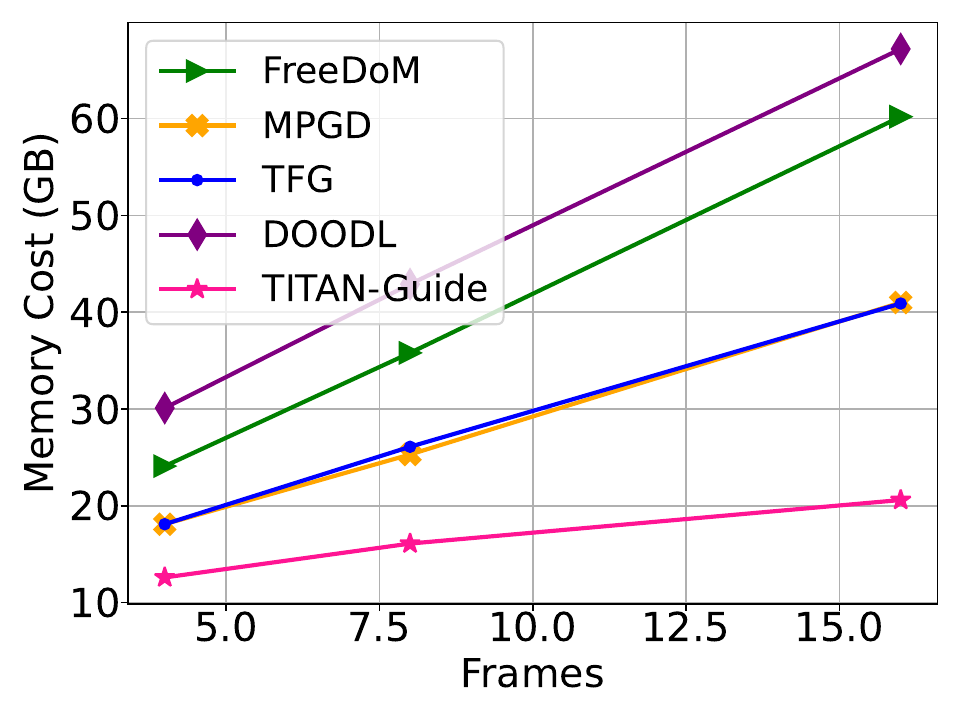}} 
 \subfloat[384x384 Resolution]{
        \includegraphics[width=0.222\textwidth]{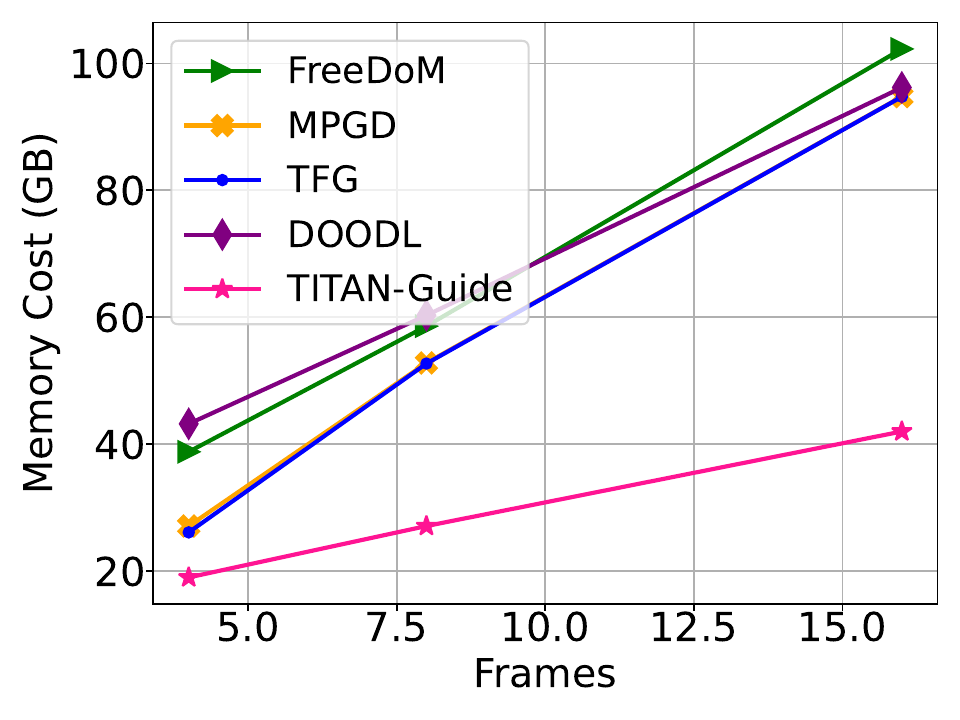}}  
\caption{Memory consumption (in GB) for various classifier-guidance methods on different video resolutions. We used the audio-visual alignnment task using Animatediff for measurement.}
\vspace{-0.28cm}
     \label{fig:memory_measure}    
\end{figure}

\paragraph{Higher resolution.} Table~\ref{table:vgg_highres} shows further improvements with higher video resolutions using AnimateDiff. By increasing resolution from 256$\times$256 to 384$\times$384, we observed performance gains across all metrics, with more generated details. TITAN-Guide used only ~42GB of memory, while the baselines exceeded 90GB, exceeding the NVIDIA H100’s capacity. Using CogVideoX with 256$\times$384 resolution, TITAN-Guide consumed around 39GB GPU memory.


\vspace{-0.25cm}
\paragraph{Memory measurement.} 
Using AnimateDiff~\cite{guo2023animatediff}, we compare memory consumption at different resolutions and frame counts, as shown in Fig. \ref{fig:memory_measure} using NVIDIA H200. For 256$\times$256 resolution, TITAN-Guide significantly reduced memory usage compared to the baselines~\cite{he2024mpgd, wallace2023doodl, ye2024tfg, yu2023freedom}, which struggled to fit on a 24GB consumer GPU. TITAN-Guide was more efficient, reducing about half the memory for T2V guidance tasks. 
\section{Conclusions}
We presented TITAN-Guide, a novel diffusion guidance method designed for memory efficiency, enabling use in memory-intensive text-to-video diffusion models. TITAN-Guide employs forward gradient descents with different types of gradient guesses and achieves competitive results in guided diffusion tasks while using less memory than previous classifier-guidance methods.

\vspace{0.2cm}
\noindent\textbf{Acknowledgements}
We gratefully acknowledge the support of all those who contributed to this research. We thank  Naoki Murata and Koichi Saito for their valuable feedback on refining and improving this work. 
{
    \small
    \bibliographystyle{ieeenat_fullname}
    \bibliography{main}
}
\newpage

\section{The Details of Forward Gradient Descents}
In this section, we provide detailed explanation on using forward AD to estimate gradients. Let $f: \mathbb{R}^m \xrightarrow{} \mathbb{R}^n$. The directional gradient along $\Vec{V}$ evaluated at $\Vec{X}$ can be defined as:
\begin{equation}
    f^\prime(\Vec{X}) = \lim_{\delta \to 0} \frac{f(\Vec{X}+\delta\Vec{V})-f(\Vec{X})}{\delta}.
\end{equation}
The forward gradient method approximates using this forward AD method with only a single forward run involved during one iteration.  Initially, the estimation of a directional gradient or a gradient guess $\Vec{V}$ is initialized with standard basis vectors. In forward gradient descents~\cite{baydin2022fwdgrad}, the gradient guess $\Vec{V}$can be initialized with random Gaussian noise, allowing for a single forward-mode run instead of multiple runs. In the modern deep learning framework (\eg, PyTorch~\cite{paszke2017automatic}), this operation can be performed with \texttt{torch.func.jvp}. 

\section{Experimental Settings}
\noindent{\textbf{Text-to-Video (T2V) Models.}} In all experiments, we employ AnimateDiff~\cite{guo2023animatediff} with epiCRealism~\footnote{\url{https://huggingface.co/emilianJR/epiCRealism}} as the base text-to-image model to generate 16 frames 8fps. We set the denoising process to 20 iterations, as we found this to be sufficient for generating clear videos. Because we work on computing gradients and updating the latent noise, we opt to use FP32 that is more stable in optimization.

\noindent{\textbf{Hyperparameters.}} For all experiments, we set $\lambda_t=0.1$. For the sampled gradient guess technique, we set the number of sampled frames $\mathcal{F}=2$, where we found that this number does not impact on significant memory usage. 

\noindent{\textbf{Toy experiment.}} In this experiment, we aim to demonstrate the effectiveness of gradient approximation using forward gradient descent, which offers improved memory efficiency. The toy dataset consists of moon-shaped data, and we employ a diffusion model to fit the generated dataset. Our diffusion model comprises two layers of neural networks, each with a hidden dimension of 64. Each layer includes a linear transformation followed by a ReLU activation function. For the diffusion process, we use Denoising Diffusion Probabilistic Models (DDPM)~\cite{ho2020denoising} as the foundational equation to fit the toy dataset.

\paragraph{Details of guidance tasks.} 
Given a text prompt, our guidance diffusion process incorporates three essential tasks: aesthetic score guidance, style guidance, and audio-video alignment.
For \textbf{aesthetic score guidance}, we build upon the classifier model of DOVER~\cite{wu2023dover}, which is designed to enhance the semantic and compositional alignment of generated content. 
In \textbf{style guidance}, we introduce a reference image that serves as a target for guiding the visual style of the generated videos. This approach ensures that the output retains a consistent artistic or cinematic look. To achieve this, we utilize Style-CLIP~\cite{patashnik2021styleclip}, a powerful model that enables classifier-based guidance for style adaptation. 
For \textbf{multi-modal alignment} (\ie, audio-video alignment), we employ ImageBind~\cite{girdhar2023imagebind}, which facilitates the alignment of target audio representations with the generated videos. This step is crucial for ensuring that the visual elements correspond appropriately to the accompanying sounds, creating a seamless and natural viewing experience. By leveraging ImageBind, we enhance the coherence between auditory and visual modalities, improving the overall realism and immersive quality of the generated videos.
Another task is \textbf{frame interpolation} with the first and end frames are provided and the model needs to interpolate the frames initialized by the latent noise.
For aesthetic score and style guidance tasks, we make use of TFG-1000Prompt dataset~\cite{ye2024tfg} and for multi-modal alignment and frame interpolation, we use VGG-Sound~\cite{chen20vggsound}.
To effectively optimize all these tasks, we design our guidance mechanisms using a loss function based on the cosine similarity between the feature representations of the generated video and the target. This ensures that the generated output closely aligns with the intended aesthetic, style, and multi-modal synchronization. For all experiments, we use 16 frames with 8 fps.

\noindent{\textbf{Frame interpolation experiments.}} Frame interpolation includes filling in missing segments of a video. In our experiment, we retain the first and last frames while reconstructing the intermediate frames. The objective is to generate a video that maintains the continuity and overall quality of the target videos. The task in the guided diffusion process is to produce a clean video: 
\begin{equation}
    \max_{\Vec{X}_0} p(\Vec{X}_0) = \max_{\Vec{X_0}} \exp(- \| \mathcal{A}(\Vec{X}_0) - \Vec{y}\|),
\end{equation}
where $\mathcal{A}(.)$ is an operator for interpolating the missing segments and $\Vec{y}$ is an original video and $\Vec{X}_0$ is the corrupted visual input.






\section{Additional Results on Video Generation}
In this section, we present our qualitative results, with additional videos available in the supplementary material.
We also provide our generated video samples at our project page\footnote{Project page: \url{https://titanguide.github.io/}}.

\begin{figure}[t]
    \centering
    \includegraphics[width=.48\textwidth]{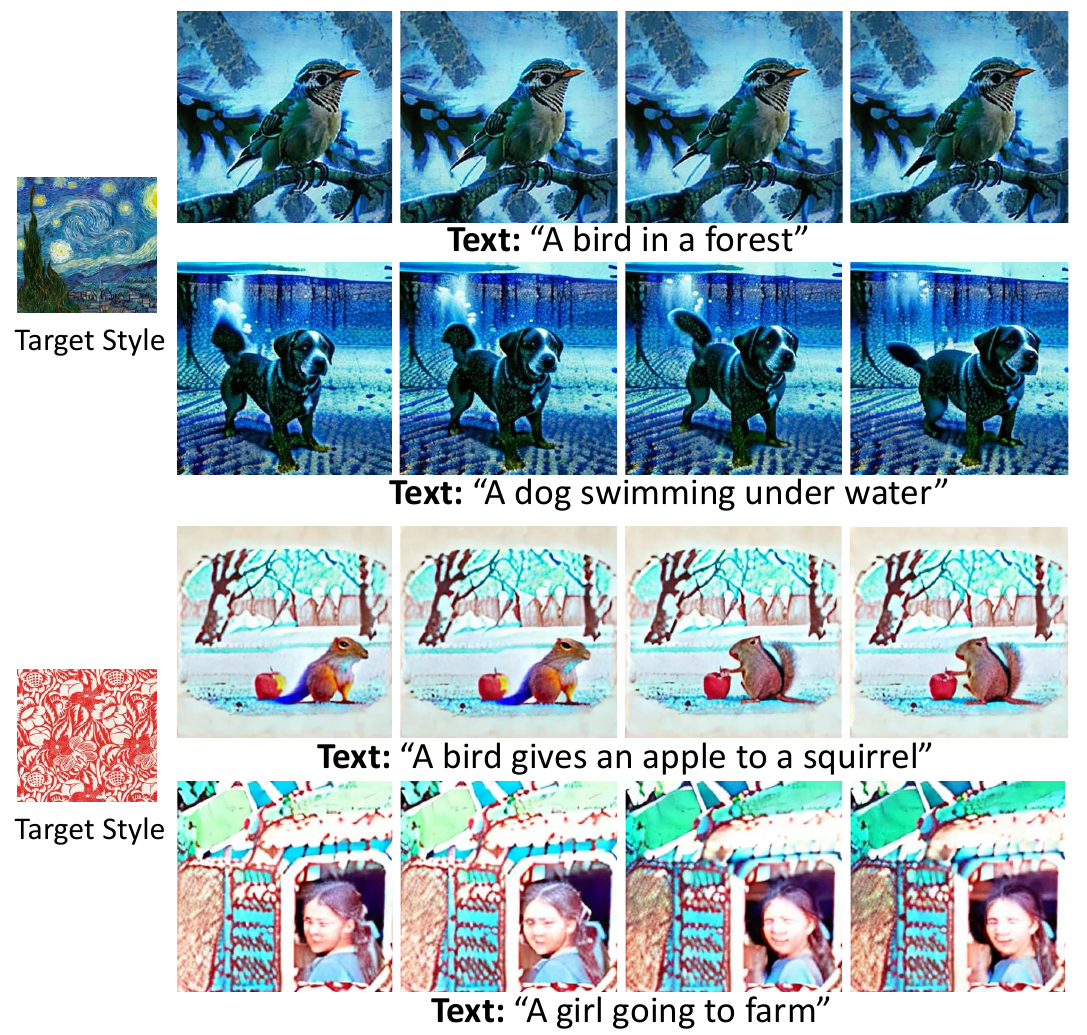}
    \caption{ Qualitative results of TITAN-Guide on style guidance.}
    \label{fig:style_additional_result}
\end{figure}

\begin{figure}[t]
    \centering
    \includegraphics[width=.48\textwidth]{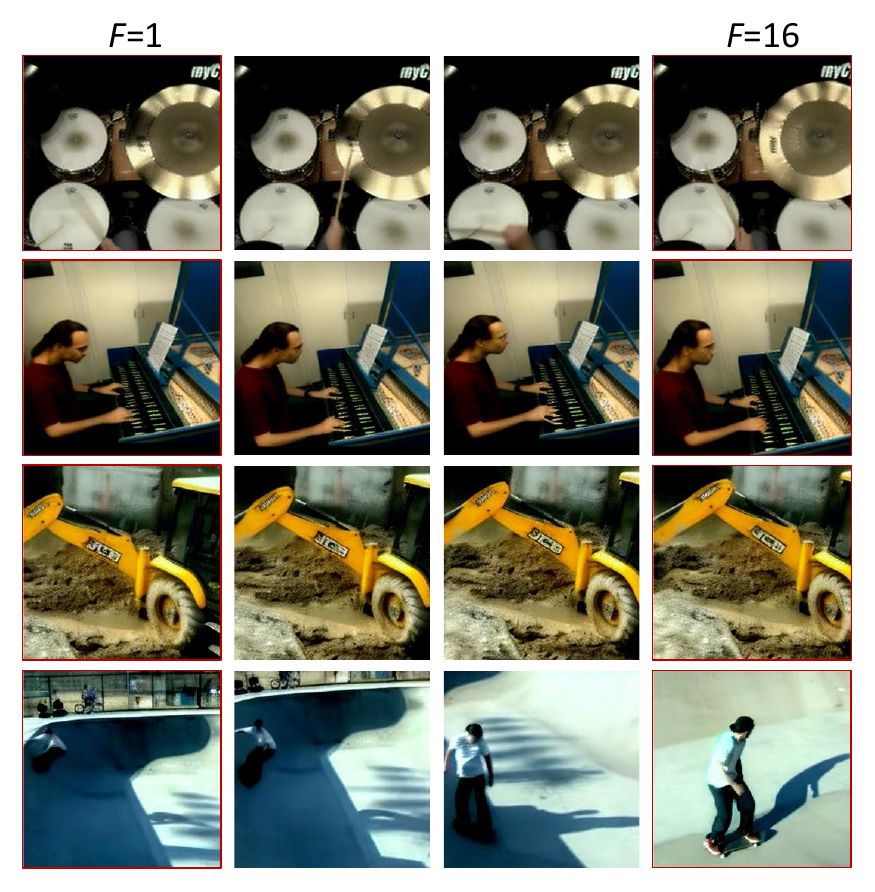}
    \caption{ Qualitative results of TITAN-Guide on frame interpolation. The given frames are indicated by \textcolor{red}{red} and interpolated frames are in the between these two frames.}
    \label{fig:interpolation_result}
\end{figure} 

\begin{figure*}[t]
    \centering
    \includegraphics[width=.98\textwidth]{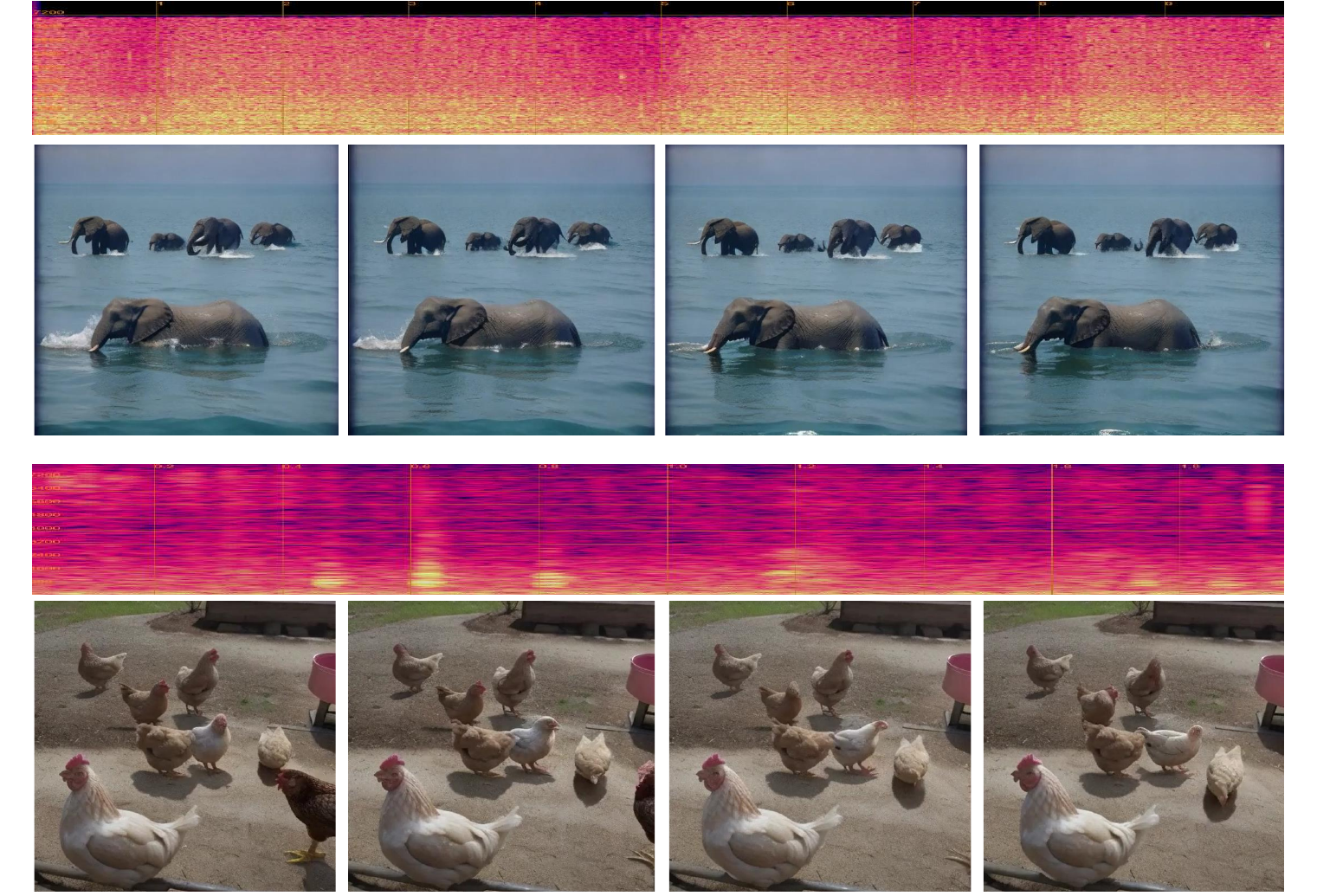}
    \caption{ Qualitative results of TITAN-Guide for audio-video alignment at 384$\times$384 resolution. Top: Elephants in water accompanied by water surface sounds. Bottom: Chickens clucking in sync with the corresponding clucking sound.}
    \label{fig:highres_result1}
\end{figure*}

\begin{figure*}[t]
    \centering
    \includegraphics[width=.98\textwidth]{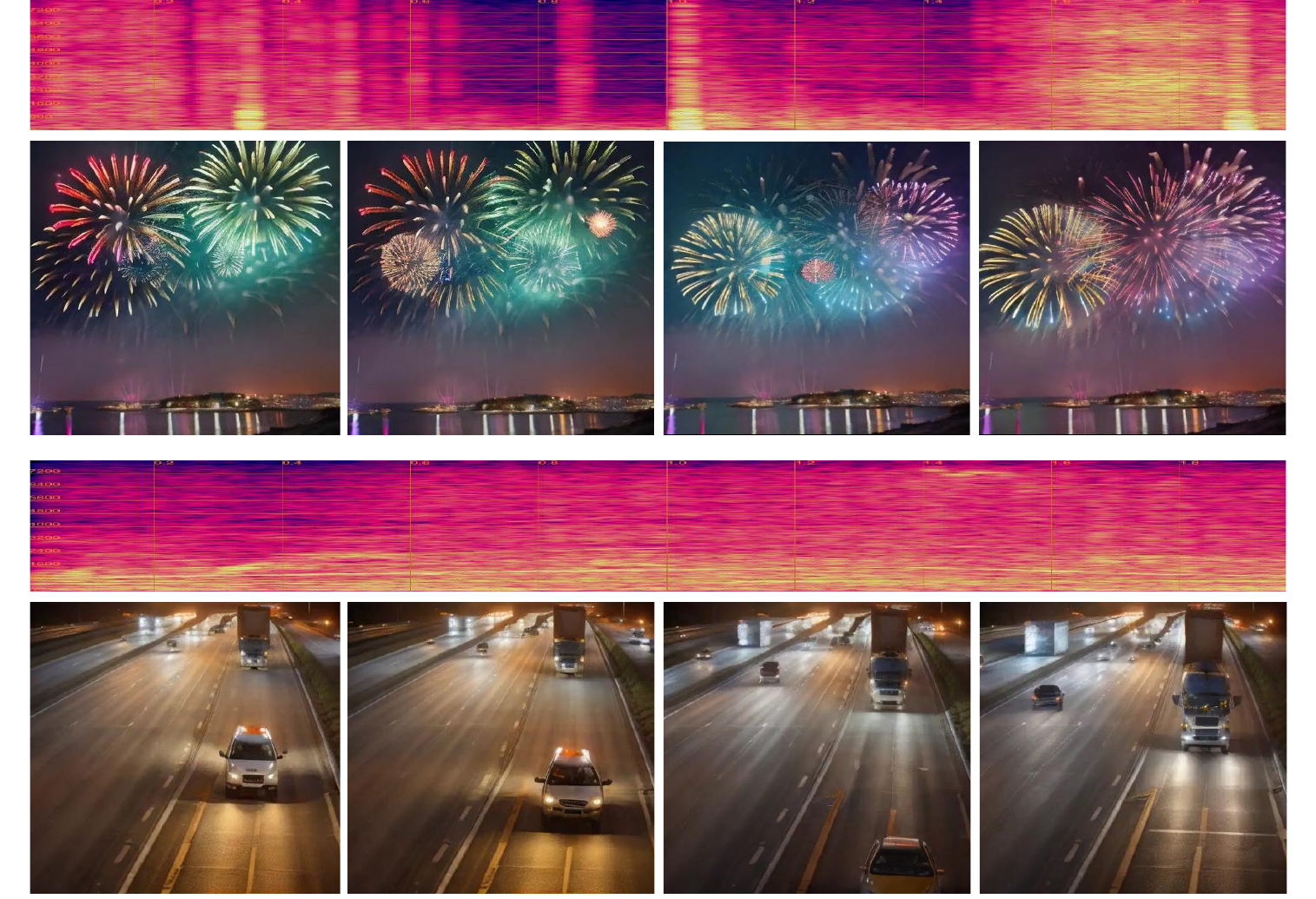}
    \caption{Qualitative results of TITAN-Guide for audio-video alignment at 384$\times$384 resolution. Top: City fireworks synchronized with firework sounds. Bottom: Highway cars accompanied by accelerating engine sounds.}
    \label{fig:highres_result2}
\end{figure*}

\paragraph{Qualitative results.} We present qualitative results illustrating the generated samples in ig.\ref{fig:style_additional_result} and\ref{fig:interpolation_result}. In the frame interpolation task, the model is given only the first and last frames of a video and have to generate the missing intermediate frames. As shown in Fig.\ref{fig:interpolation_result}, our proposed method effectively completes this task, demonstrating the capability in temporal consistency. Additionally, for style guidance, we showcase two distinct visual styles in Fig.\ref{fig:style_additional_result}, further highlighting the efficacy of our approach.  We also provide generated video results of 384$\times$384 resolution in Fig.~\ref{fig:highres_result1} and~\ref{fig:highres_result2}.

\noindent \textbf{Time and memory consumption}. Below, we discuss the time require to process a video on 16 frames with 256$\times$256 resolution. We use H200 to evaluate the processing time. TITAN-Guide requires 2 minutes to perform guidance in diffusion models, and DOODL~\cite{wallace2023doodl} requires a similar time with ours. As we know the other methods \eg, TFG~\cite{ye2024tfg}, MPGD~\cite{he2024mpgd}, FreeDoM~\cite{yu2023freedom}, Seeing and Hearing~\cite{xing24seeing} do not traverse to $t=0$, they require about 40 seconds for video generation. 

\section{Additional Results on Image Generation}
We also demonstrate that our proposed approach has the potential to be effectively applied to the image diffusion models~\cite{ho2020denoising,song2020ddim}. However, since our primary focus is text-to-video tasks, we do not explore this aspect in depth but provide evidence of its applicability.

\paragraph{Settings.}
 In all image generation experiments, we use 256$\times$256 image resolution.
We use three different tasks: 1) Super resolution, 2) CelebA (guided by gender and age specification), and deblurring. Following~\cite{ye2024tfg}, for super resolution, and deblurring tasks, we use the CAT-DDPM diffusion model trained on the CAT dataset~\cite{Elson2007AsirraAC}. While, we use CelebA-DDPM trained on the CelebA dataset~\cite{liu2015faceattributes} for gender\footnote{Gender: \url{https://huggingface.co/rizvandwiki/gender-classification}} and age\footnote{Age: \url{https://huggingface.co/londe33/hair_v0}} guidance. 
\paragraph{Evaluation.} For the experiment in image domain, we evaluate based on Learned Perceptual
Image Patch Similarity (LPIPS) and Fréchet Image Distance (FID) to assess the quality of generated images. For the CelebA (gender and age) task, we measure classification accuracy (Acc.) and Kernel Inception Distance (KID) to  to assess fidelity of generated samples. For all experiments, we generate 256 images to evaluate the effectiveness of our proposed method.

 \begin{table}[h]
 \centering
    \resizebox{0.489\textwidth}{!}{
    \Large\addtolength{\tabcolsep}{.1pt}
\begin{tabular}{lcccccc}
\toprule
\multirow{2}{*}{Method}     &\multicolumn{2}{c}{Super Resolution} &\multicolumn{2}{c}{CelebA} &\multicolumn{2}{c}{Deblurring}\\    
 &LPIPS$\downarrow$ &FID$\downarrow$ &Acc.$\uparrow$ &KID$\downarrow$ &LPIPS$\downarrow$ &FID$\downarrow$ \\
\hline
FreeDoM~\cite{yu2023freedom}            & 0.191& 74.5& 68.7                                                                  & 3.89     & 0.245 &87.4                         \\

MPGD~\cite{he2024mpgd}                & 0.283 & 82.0 & 68.6                                                                    & 4.79    & 0.177 & 69.3                        \\
TFG~\cite{ye2024tfg}              & 0.190 & \textbf{65.9} & 75.2                                                                  & 3.86     & \textbf{0.150} & 64.5                          \\
  \rowcolor{aliceblue} \;TITAN-Guide (ours) & \textbf{0.180} & 75.9                                                              & \textbf{77.2} & \textbf{3.61}    &  0.222                                                                 &    \textbf{63.5}                      \\  
 \bottomrule
 \bottomrule
\end{tabular}
}
\caption{Evaluation results across various metrics for assessing the quality of generated images.}
\label{table:imagedomain}
\end{table}

\paragraph{Results.} Table~\ref{table:imagedomain} shows the results of the three tasks. In this experiment, we use TITAN-Guide exclusively with sampled gradient guesses. Our observations indicate that TITAN-Guide outperforms previous methods in most tasks and metrics. Additionally, the image generation results demonstrate its effectiveness in image guidance tasks.

\end{document}